\documentclass[runningheads]{llncs}

\usepackage{eccv}

\usepackage{eccvabbrv}
\usepackage{multirow}
\usepackage{colortbl}
\usepackage{graphicx}
\usepackage{booktabs}
\usepackage{bbding}
\usepackage[accsupp]{axessibility}  
\usepackage{pifont}
\usepackage{algorithm}
\usepackage{algorithmic}
\newcommand{\cmark}{\ding{51}}%
\usepackage[pagebackref,breaklinks,colorlinks,citecolor=eccvblue]{hyperref}
\usepackage[width=122mm,left=12mm,paperwidth=146mm,height=193mm,top=12mm,paperheight=217mm]{geometry}

\usepackage{orcidlink}

\begin{document}

\title{Exploring Phrase-Level Grounding with Text-to-Image Diffusion Model} 

\titlerunning{Exploring Phrase-Level Grounding with
Text-to-Image Diffusion Model}

\author{Danni Yang\inst{1} \and
Ruohan Dong\inst{1} \and
Jiayi Ji\inst{1} \and
Yiwei Ma\inst{1} \and
Haowei Wang\inst{1,2} \and \\
Xiaoshuai Sun\inst{1}\thanks{The corresponding author} \and
Rongrong Ji\inst{1} }

\authorrunning{D. Yang et al.}

\institute{Key Laboratory of Multimedia Trusted
Perception and Efficient Computing, Ministry of Education
of China, School of Informatics, Xiamen University \and
Youtu Lab, Tencent, Shanghai, China \\
\email{\{yangdanni,dongruohan\}@stu.xmu.edu.cn},
\email{jjyxmu@gmail.com}, \\
\email{\{yiweima,wanghaowei\}@stu.xmu.edu.cn},
\email{\{xssun,rrji\}@xmu.edu.cn}}


\maketitle

\begin{abstract}
Recently, diffusion models have increasingly demonstrated their capabilities in vision understanding. By leveraging prompt-based learning to construct sentences, these models have shown proficiency in classification and visual grounding tasks. However, existing approaches primarily showcase their ability to perform sentence-level localization, leaving the potential for leveraging contextual information for phrase-level understanding largely unexplored. In this paper, we utilize Panoptic Narrative Grounding (PNG) as a proxy task to investigate this capability further. PNG aims to segment object instances mentioned by multiple noun phrases within a given narrative text. Specifically, we introduce the DiffPNG framework, a straightforward yet effective approach that fully capitalizes on the diffusion's architecture for segmentation by decomposing the process into a sequence of localization, segmentation, and refinement steps. The framework initially identifies anchor points using cross-attention mechanisms and subsequently performs segmentation with self-attention to achieve zero-shot PNG. Moreover, we introduce a refinement module based on SAM to enhance the quality of the segmentation masks. Our extensive experiments on the PNG dataset demonstrate that DiffPNG achieves strong performance in the zero-shot PNG task setting, conclusively proving the diffusion model's capability for context-aware, phrase-level understanding. Source code is available at \href{https://github.com/nini0919/DiffPNG}{https://github.com/nini0919/DiffPNG}.
  \keywords{ Diffusion Model \and Panoptic Narrative Grounding \and Zero-Shot Learning}
\end{abstract}

\begin{figure*}[t] 
\centering 
\includegraphics[width=1\columnwidth]{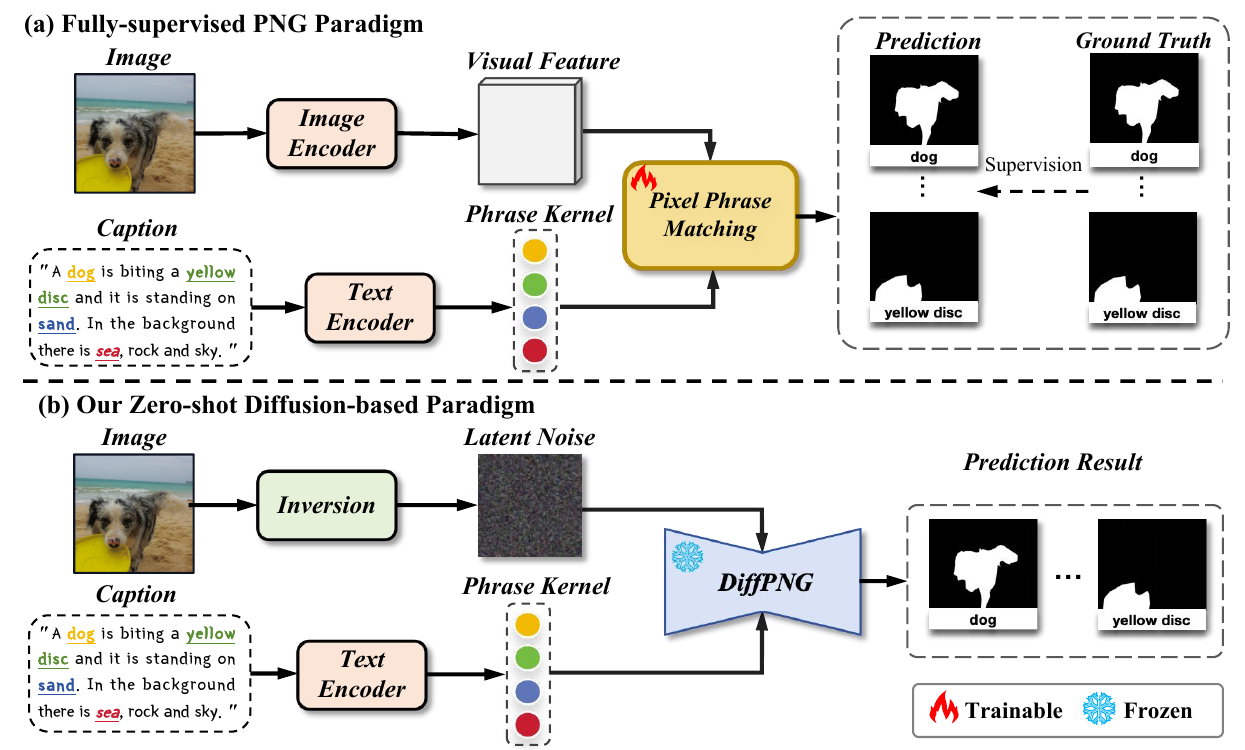}  
\vspace{-0.8em}
\caption{A comparison between the previous fully-supervised PNG paradigm with our proposed Zero-Shot Diffusion-based Paradigm. Motivated by the strong image-text alignment of text-to-image diffusion models, we employ these generative models in our PNG task through a zero-shot manner. This is aimed at exploring the ability of diffusion models to perform phrase-level grounding.} 
\label{fig:fig1}
\vspace{-2em}
\end{figure*}
\section{Introduction}
\label{sec:intro}

The recent development of large-scale text-to-image diffusion models~\cite{ramesh2022hierarchical,rombach2022high,saharia2022photorealistic}, notably Stable Diffusion~\cite{rombach2022high}, has demonstrated remarkable capabilities in producing realistic and varied imagery. These models have been trained on extensive text-image datasets like LAION-5B~\cite{schuhmann2022laion}, enabling them to understand the complex relationships between text descriptions and visual elements, and thus, achieve outstanding vision-language alignment.\par
This advancement has led researchers to explore the capabilities of diffusion models beyond mere image generation, applying them to vision understanding tasks. Early attempts~\cite{li2023your} utilized diffusion models as zero-shot classifiers through the use of specific prompts, opening the door to their application in perception tasks without finetuning. Subsequent research~\cite{li2023open,chen2023diffusiondet,ma2023diffusionseg,wu2023diffumask,nguyen2024dataset} extended these capabilities to detection and segmentation tasks, highlighting the models' strong potential for accurate localization. Further applications in referring expression comprehension/segmentation~\cite{liu2023vgdiffzero,ni2023ref} have solidified the belief in the significant potential of this approach. The success of these models in various tasks underscores their ability for language-driven visual perception. However, these tasks have primarily showcased the models' ability to understand visual content at the sentence level, depending on complete sentences or those crafted using prompt techniques.  The exploration into their capability for context-aware phrase-level understanding remains relatively unexplored.

Acknowledging this gap, we identify the need for a task that can further this exploration. The Panoptic Narrative Grounding (PNG) task~\cite{gonzalez2021panoptic,wang2023towards,ding2022ppmn,wang2023nice,lin2023context,hui2023enriching,gonzalez2023piglet,yang2023semi}, which has seen considerable advancement as a pixel-level multimodal challenge, fits this requirement perfectly. PNG aims to generate pixel-level masks for each noun phrase within a long narrative sentence, as shown in Fig.~\ref{fig:fig1}(a), offering a more fine-grained comprehension compared to other cross-modal tasks, such as visual question answering~\cite{zhou2015simple,shih2016look,wu2017visual,kafle2017visual} and image captioning~\cite{xu2015show,vinyals2016show,cornia2020meshed,pan2020x,luo2021dual}. This task serves as an ideal platform to delve into the phrase-level understanding capabilities of diffusion models, moving beyond sentence-level understanding to more granular, context-aware processing.

In this paper, we introduce DiffPNG, a straightforward yet innovative zero-shot learning framework for Panoptic Narrative Grounding (PNG) that utilizes text-to-image diffusion models, as illustrated in Fig.~\ref{fig:fig1}(b). Specifically, DiffPNG views the PNG task as a localization-segmentation-refinement paradigm, primarily facilitated by our proposed \emph{Locate-to-Segment Processor} (LSP) and \emph{SAM-based Mask Refinement} (SMR) modules. Within LSP, we fully leverage the characteristics of cross-attention and self-attention, creatively decoupling localization and segmentation. Initially, we utilize cross-attention during the reverse diffusion process to identify the location of each instance and obtain anchors. Subsequently, using self-attention, we aggregate these anchors to generate the segmentation mask. To further improve the coherence of the masks' edges, we develop SMR, which utilizes SAM~\cite{kirillov2023segment} to refine our masks. We devise a strategy to align the candidate masks generated by SAM with our segmentation outcomes, thus producing a more refined output. Extensive experiments conducted on the PNG benchmark dataset demonstrate that the proposed DiffPNG achieves exceptional zero-shot performance, proving the diffusion model's capability for context-aware, phrase-level understanding.

\begin{itemize}
    \item To investigate the phrase-level vision understanding capabilities of diffusion models, we reformulate the Panoptic Narrative Grounding task as a zero-shot problem that begins with initial localization, proceeds to segmentation, and concludes with refinement.
    \item We present the LSP module, which utilizes cross-attention to pinpoint anchor pixels and self-attention for the aggregation and segmentation processes. Additionally, we develop the SMR strategy, harnessing the capabilities of SAM to significantly refine the segmentation mask.
    \item The proposed framework has achieved superior zero-shot performance on the PNG benchmark dataset, conclusively demonstrating the capability of diffusion models to understand visual content at the phrase level.
\end{itemize}

\section{Related Work}
\subsection{Panoptic Narrative Grounding}
Panoptic Narrative Grounding (PNG) aims to segment the objects and regions corresponding to nouns in a long narrative sentence within a given image, enabling more sophisticated scene understanding. Similar to other multi-modal visual segmentation tasks~\cite{yang2022lavt,qian2024x}, the core challenge of the PNG task lies in the understanding of both textual and visual semantics. González et al.\cite{gonzalez2021panoptic}  pioneered this novel task by creating a benchmark with a new standard dataset and evaluation methodology, while also proposing a fully-supervised baseline method as a starting point for future work. The baseline method follows a two-stage approach: a pre-trained panoptic segmentation model extracts a large number of segmentation region proposals, and a Region-Phrase matching scoring module evaluates candidates to determine the segmentation masks for each noun phrase. This process heavily relies on the results of the first stage and lacks elegance.
To tackle this issue, Ding et al.~\cite{ding2022ppmn} proposed a one-stage Pixel-Phrase Matching Network that directly matches each phrase to its corresponding pixels and outputs panoptic segmentation. Additionally, Wang et al. proposed EPNG~\cite{wang2023towards} to tackle the high computational and spatial costs in the segmentation stage, particularly focusing on real-time applications. Furthermore, Yang et al. introduced SS-PNG~\cite{yang2023semi} based on semi-supervised learning, which improves PNG's progress in the face of expensive annotation costs. 
\subsection{Generative Model for Visual Segmentation}
Generative models, as a class of machine learning models, aim to learn the probability distribution of data to generate new data samples. 
In recent years, there has been extensive research in the field of generative models, including Generative Adversarial Networks (GANs)~\cite{brock2018large,esser2021taming,karras2019style,karras2020analyzing,zhu2017unpaired}, and the Diffusion Models~\cite{dockhorn2021score,ho2020denoising,jolicoeur2020adversarial,nichol2021improved,sohl2015deep,song2020denoising,song2019generative,song2020improved,song2020score}.
Especially, diffusion models has achieved significant success in text-to-image generation systems, such as DALL-E~\cite{ramesh2022hierarchical}, Imagen~\cite{saharia2022photorealistic}, and Stable Diffusion~\cite{rombach2022high}. Due to the robust abilities of these generative models in aligning images with text and comprehending their relationship, recent researchers have explored their potential in perceptual tasks, including dense prediction tasks such as semantic segmentation~\cite{baranchuk2021label,galeev2021learning,li2022bigdatasetgan,meng2021sdedit,tritrong2021repurposing,zhang2021datasetgan} to generate pixel-level segmentation results. 
Recent works on segmentation using GANs primarily focus on the segmentation of prominent objects~\cite{luc2016semantic,lutz2018alphagan,chen2017no,hoffman2016fcns,tsai2018learning,ehsani2018segan}, which have yet to capture the intricate complexities present in real-world scenes.
Surprisingly, Due to the powerful image-text comprehension capabilities of diffusion models, diffusion-based visual segmentation has made significant progress~\cite{baranchuk2021label,li2023open,wu2023diffumask,nguyen2024dataset,tian2023diffuse,ma2023diffusionseg,karazija2023diffusion}, fully leveraging the intrinsic value of diffusion models.
These works can be roughly categorized into three types.
The first type involves generating synthetic datasets using diffusion models to reduce annotation cost~\cite{wu2023diffumask,nguyen2024dataset}. 
The second involves utilizing the U-Net structure of pre-trained diffusion models for feature extraction~\cite{shao2023semi,xu2023open,pnvr2023ld,li2023guiding,tang2022daam}. 
The third type involves using cross-attention and self-attention mechanisms to assist the model in capturing contextual information within images~\cite{tian2023diffuse,ma2023diffusionseg,tian2023diffuse}. 
The emergence of these methods has made segmentation task training more efficient and has reduced the reliance on pixel-wise labels. Inspired by the aforementioned work, we aim to propose a zero-shot diffusion-based method tailored specifically for the PNG task.

\subsection{Zero-Shot Segmentation}
Zero-shot segmentation refers to segmenting objects on unseen samples. There are mainly three kinds of methods in zero-shot learning: training a large model that performs well in almost all unseen images~\cite{Wang2023Seggpt,kirillov2023segment,ren2024grounded}; utilizing models that are not specifically trained for segmentation tasks~\cite{xu2021,shin2022reco,zhou2023zegclip,ding2022decoupling,xu2022simple}; unsupervised or self-supervised learning~\cite{caron2021emerging,cho2021picie,feng2023network,li2023acseg,shin2023namedmask}. Large segmentation model like SAM~\cite{kirillov2023segment} shows impressive zero-shot performance on various visual tasks. Although this kind of approach can achieve relatively accurate segmentation results, it requires a significant amount of time to create such a high-quality dataset.
Reco~\cite{shin2022reco} performs zero-shot segmentation by retrieving and segmenting based on a pre-trained image encoder and CLIP.
Self-supervised learning method DINO~\cite{caron2021emerging} explicitly contains the scene layout and object boundaries, which can easily transfer to downstream object segmentation. Global-Local CLIP~\cite{yu2023zero} proposed zero-shot transfer from pre-trained CLIP to RES.

\section{Method}
In this section, we introduce our simple yet effective DiffPNG network for Panoptic Narrative Grounding, which utilizes a pre-trained Stable Diffusion model to generate high-quality segmentation masks. We will briefly review diffusion model and introduce our problem definition in Sec.~\ref{sect:pre}
Then we will present our architecture and inference details, including Feature Extraction(Sec.~\ref{sect:feature}), Locate-to-Segment Processor(Sec.~\ref{sect:csai}), Subject-Focused Feature Aggregator (Sec.~\ref{sffa}), SAM-based Mask Refinement(Sec.~\ref{smr}).
\begin{figure*}[t] 
\centering 
\includegraphics[width=1\columnwidth]{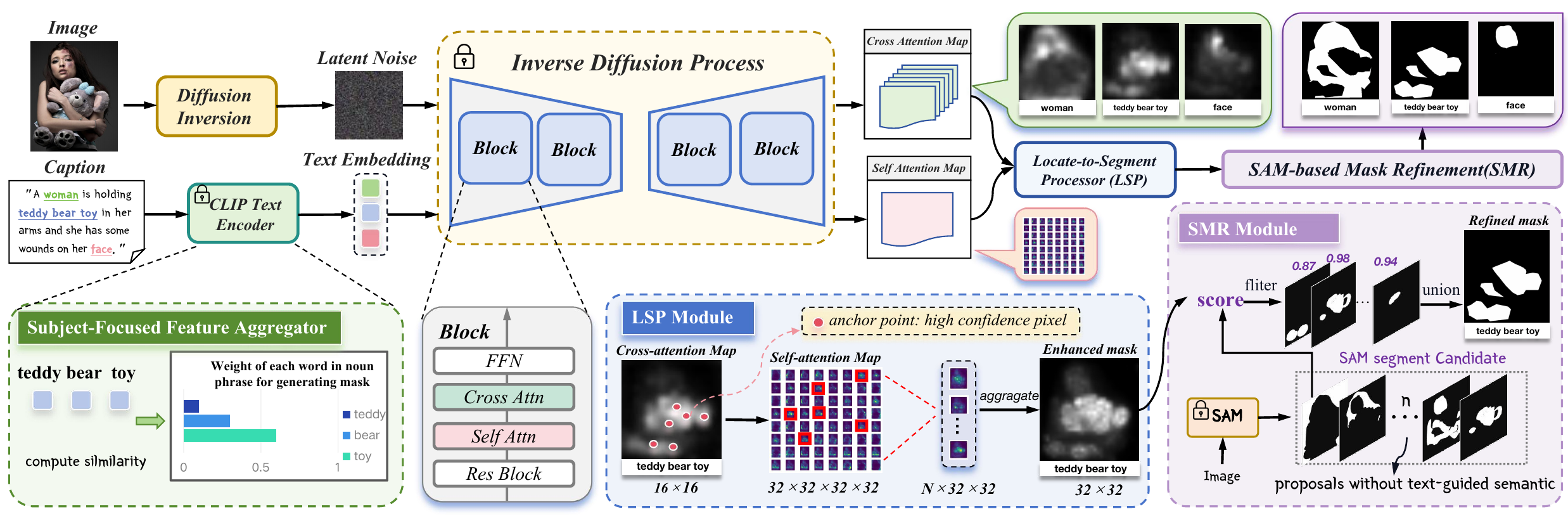}  
\vspace{-1.8em}
\caption{Overview of the proposed DiffPNG framework, of which all components, \emph{i.e.}, Feature Extraction, Locate-to-Segment Processor, Subject-Focused Feature Aggregator, SAM-based Mask Refinement.} 
\label{fig:overview}
\vspace{-1em}
\end{figure*}

\subsection{Preliminaries}
\label{sect:pre}
\noindent\textbf{Diffusion Model.}
Diffusion models represent a novel class of generative models, showcasing remarkable success in text-to-image generation systems~\cite{ramesh2022hierarchical,rombach2022high,saharia2022photorealistic}. Among these, Stable Diffusion~\cite{rombach2022high} is a type of latent diffusion model that implements diffusion processes in latent space rather than in pixel space. Stable diffusion models utilize an encoder-decoder framework. The encoder $E$ compresses the input image $x \in \mathbb{R}^{H \times W \times 3}$ into a smaller spatial latent space $z \in \mathbb{R}^{h \times w \times c}$, expressed as $z=E(x)$, while the decoder $D$ reconstructs the image from $z$, written as $\tilde{x}=D(z)$.
The diffusion process consists of a forward process and a reverse process. During the forward process, a small amount of Gaussian noise is gradually added at each time step until the image becomes an isotropic Gaussian noise image.  During the reverse process, diffusion models are trained to gradually remove Gaussian noise to restore the original clear image as follows:
\begin{equation}
z_{t-1}=\sqrt{\frac{\alpha_{t-1}}{\alpha_t}} z_t+\left(\sqrt{\frac{1}{\alpha_{t-1}}-1}-\sqrt{\frac{1}{\alpha_t}-1}\right) \cdot \varepsilon_\theta\left(z_t, t, \mathcal{C}\right),
\end{equation}
where $z_t$ is $t$-th step noisy latent, $\beta_1,\ldots,\beta_T$ is variance schedule and $\alpha_t = 1-\beta_t$, $\mathcal{C}$ is conditional input, $\varepsilon_\theta$ denotes the denoising U-Net.\par
 All diffusion processes take place in the latent space, implemented through a U-Net~\cite{ronneberger2015u} architecture with attention layers. This paper focuses on cross-attention and self-attention in the U-Net architecture during the inverse process to construct our paradigm.\par
\noindent\textbf{Problem Definition.}
In our setting, our goal is to utilize Panoptic Narrative Grounding (PNG) as a proxy task to investigate the phrase-level understanding capability of diffusion model. Specially, given an image $\mathcal{I} \in \mathbb{R}^{H \times W \times C_v}$ and corresponding long narrative text $\mathcal{T}\in \mathbb{R}^{N_t \times C_t}$ from the PNG validation set, our model aims to output the segmentation masks $\mathcal{M} \in \mathbb{R}^{N_t' \times H \times W}$, where $H$, $W$, $C_v$ represent the height, width, channel of image $\mathcal{I}$, and $N_t$, $N_t'$ represents the maximum and the exact number of noun phrases, and $C_v$ denotes the channel of the noun phrases. To this end, we use $\Phi_\mathcal{D}$ to denote our DiffPNG framework, and the goal is to accomplish zero-shot inference, as outlined below:
\begin{equation}
\mathcal{M} = \Phi_{\mathcal{D}} \left( \mathcal{I}, \mathcal{T} \right)\in\{0,1\}^{N_t' \times H \times W}, \mathcal{I} \in \mathbb{R}^{H \times W \times C_v}, \mathcal{T} \in \mathbb{R}^{N_t \times C_t}.
\end{equation}
In our zero-shot Panoptic Narrative Grounding settings, our model will not access any training data, including images, narrative texts, and mask annotations. Our proposed framework is depicted in Fig.~\ref{fig:overview}. In the following sections, we will provide a detailed introduction to each module of our framework.

\subsection{Feature Extraction}
\label{sect:feature}
In diffusion models, latent noise and text embedding are provided as inputs to perform the diffusion process. Consequently, the first step is to obtain the latent representation and text embedding.\par
\noindent\textbf{Visual Modality.} 
In fully-supervised PNG tasks, FPN~\cite{lin2017feature} with a ResNet-101~\cite{he2016deep} is usually used to extract high-dimensional features of images for fusion with text features. 
However, for tasks involving stable diffusion~\cite{rombach2022high}, where accepting latent noise variables is necessary, this approach needs adjustment. Real images from the PNG dataset must be converted into latent noise, a process typically handled by the DDPM~\cite{ho2020denoising}, which introduces noise to generate latent variables. Unfortunately, the randomness of these variables impedes accurate image reconstruction during the reverse process, making them unsuitable for feature extraction.
To maintain consistency in reconstruction, we employ the Null-text Inversion~\cite{mokady2023null}, with DDIM inversion as its core, ensuring the reconstructed images closely match the originals and remain relevant to their descriptions. This allows the latent noise to be effectively used for further segmentation tasks.

Given $T$ sampling steps, DDIM inversion outputs noise latent $\{z^*_{t}\}^T_{t=1}$, Null-text inversion outputs latent $\{\overline{z}_{t}\}^T_{t=1}$, the initial T step noisy latent $\overline{z}_T$ is equal to $z_T$. To prevent a significant change in the reconstructed image, we minimize Null-text inversion loss for time $t = T,...,1$ as:

\begin{equation}
    \underset{\varnothing_t}{min}~\mathcal{L}^{inv}_{t-1}= \|{z}^*_{t-1} - f_{t-1}(\overline{z}_t,\varnothing_t,\mathcal{C})\|,    
\end{equation}
After N iterations optimization for unconditional embedding at time $t$, we update the inverted noisy latent by 
current noisy latent $\overline{z}_{t-1} = f_{t-1}(\overline{z}_t,\varnothing_t,\mathcal{C})$, where $f_{t-1}(\cdot)$ denotes a function that maps $\overline{z}_t$, $\varnothing_t$, and $\mathcal{C}$ to $\overline{z}_{t-1}$. The denoising U-Net can perceive a more accurate attention map from conditional input.

\noindent\textbf{Linguistic Modality.}
In the previous fully-supervised PNG setting~\cite{gonzalez2021panoptic,ding2022ppmn,wang2023towards,wang2023nice,yang2023semi},  the “base-uncased”
version of BERT~\cite{devlin2018bert} is used as the text encoder to extract the embedding of noun phrases. In this paper, to align with the open-source diffusion model, we utilize CLIP text encoder to encode text. Yet, limited by the maximum length accepted by the CLIP text encoder, most narrative texts in PNG dataset exceed this maximum length. Therefore, we adopted a text-splitting strategy to maximize the preservation of the paragraph's contextual semantics, ensuring that each word in the text can be fully involved in computations. Specially, we split the paragraph $\mathcal{T}$ into multiple sub-paragraphs$\{\mathcal{T}_{n}\}^{N_s}_{n=1}$, with each sub-paragraph's length not exceeding the maximum length of the text encoder. By feeding sub-paragraphs into CLIP text encoder 
$\tau_\theta$ to generate text embedding $\{e^n_{c}|e^n_{c} = \tau_\theta(\mathcal{T}_n)\}^{N_s}_{n=1}$ of long narrative text.
\subsection{Locate-to-Segment Processor}
\label{sect:csai}
The inverse diffusion process effectively denoises by leveraging latent variables derived from inversion module and text embedding from CLIP text encoder. 
This process uses a U-Net network for prediction, which comprises 16 specific blocks composed of two main components: a ResNet layer and a Transformer layer. The Transformer layer employs two attention mechanisms: self-attention to capture global 
image features and cross-attention to establish connections between the visual and textual input. Formally, the self-attention map $\mathcal{A}_{self}^{n,l} \in[0,1]^{H \times W \times H \times W}$ and cross-attention map $\mathcal{A}_{cross}^{n,l} \in[0,1]^{H \times W}$ at layer $l$ for phrase n are computed as follows: 
\begin{equation}
\mathcal{A}^{n,l}_{self}=\operatorname{Softmax}\left(\frac{Q_z K_z^T}{\sqrt{d_l}}\right) \in [0,1]^{ H \times W \times  H \times W },  
\end{equation}
\begin{equation}
\mathcal{A}^{n,l}_{cross}=\operatorname{Softmax}\left(\frac{Q_z K_e^T}{\sqrt{d_l}}\right) \in [0,1]^{ H \times W}, 
\end{equation}
where $Q_z, K_z$ are $l$-th layer's noisy latent $z^l_t$, and $K_e$ is an intermediate representation of text input. $d_l$ is the embedding dimension at layer $l$.\par
In particular, cross-attention establishes a relationship between textual features and visual pixels, allowing us to extract the corresponding pixels for noun phrases. However, employing cross-attention directly as mask candidates might be somewhat crude, potentially resulting in noisy regions. Motivated by this observation, we firstly select some pixels on the cross attention map for each noun phrase, with confidence levels higher than a certain threshold $\beta$, as anchor points, which are defined as:
\begin{equation}
    \mathcal{P}^n_{anchor}=\left\{(i,j) \mid \mathcal{A}^{n,avg}_{cross}(i,j) > \beta  \right\},
\end{equation}
where $\mathcal{P}^n_{anchor}, \mathcal{}{A}^{n,avg}_{cross}$ are the anchor point set and the average cross-attention of each cross-attention layer for phrase $n$, respectively. \par
These anchor points effectively highlight the most relevant regions in the text. To obtain masks with precise semantic information, we leverage pixel-level self-attention maps with high spatial coherence to associate similar pixels, thereby achieving refined segmentation results. Specially, we aggregate each anchor pixel’s self-attention maps and apply Min-Max normalization to generate an enhanced map as follows:
\begin{equation}
\mathcal{A}^n_{enhanced}= Norm \left( \sum_{(i,j) \in \mathcal{P}^n_{anchor}}\mathcal{A}^{avg}_{self}\left(i,j\right)\right),
\label{enhance}
\end{equation}
where $\mathcal{A}^n_{enhanced}$ is the enhanced map by self-attention, $\mathcal{A}^{avg}_{self}$ is average self-attention of each selected self-attention layer. $Norm()$ is Min-Max normalization.\par
Finally, we perform a binarization on the enhanced probability score map $\mathcal{A}^n_{enhanced}$ to obtain the segmentation result as follows:
\begin{equation}
{\hat{M^n}(i,j)}= \begin{cases}0 ,& {\mathcal{A}^n_{enhanced}(i,j)} \leq \alpha \\ 1, & {\mathcal{A}^n_{enhanced}(i,j)}> \alpha \end{cases},
\label{one-hot}
\end{equation}
where $\hat{M^n}$ is the segmentation mask of the current noun phrase $n$, $\alpha$ is the binarization threshold, ranging from 0 to 1. \par
\subsection{Subject-Focused Feature Aggregator}
\label{sffa}
To better utilize cross-attention between different words within a noun phrase, we propose a Subject-Focused Feature Aggregator which is a similarity-based adaptive feature aggregation scheme. Specially, let $N_p$ denote the number of encoded tokens in the $p$-th noun phrase, represented as $v_1, v_2, \ldots, v_{N_{p}}$, where each $v_i$ represents the word embedding obtained through the CLIP text encoder. We compute the pairwise similarities between word $v_i$ and word $v_{N_p}$ using dot product, resulting in the similarity score vector $s=\left[s_1, s_2, \ldots, s_{N_p}\right]$, where defined as:
\begin{equation}
s_i=v_i \cdot v_{N_p}, \quad \text { for } i=1,2, \ldots, N_p    ,
\end{equation}
where $s_i$ represents the contribution of the word in this phrase to activating relevant regions within an image. It is worth noting that the last word in English noun phrases is always a noun, while the preceding words can be adjectives, pronouns, and numerals. Thus, we assign a weight of 1 to the last word, with the weights of the other words calculated based on their similarity scores.\par
Subsequently, we scale $s_i$ to $\left[0,1\right]$ by Softmax operator $w_i = \operatorname{Softmax}\left(s_i\right)$. We compute the aggregated attention $\mathcal{A}^n_{fuse}$ based on Eq.~\ref{enhance}  for phrase $n$, which is defined as:
\begin{equation}
\mathcal{A}^n_{fuse} = \sum^{N_p}_{i=1}w_i*\mathcal{A}^{n,i}_{enhanced},
\end{equation}
where $\mathcal{A}^n_{fuse}$ is the aggregated attention map for phrase $n$, and $\mathcal{A}^{n,i}_{enhanced}$ is the input attention map for $i$-th word in phrase $n$.\par

\subsection{SAM-based Mask Refinement}
Due to the zero-shot setting which prohibits access to the training set, the edge information in the masks generated directly by DiffPNG still needs refinement. Therefore, motivated by the strong edge segmentation capability of SAM~\cite{kirillov2023segment,yang2024sam}, we propose a refinement module called SAM-based Mask Refinement(SMR) to correct the output of DiffPNG, referred to as DiffPNG*. In detail, we feed the images into the frozen SAM without providing any prompts, utilizing SAM's "segment everything" functionality to obtain several candidate masks, which are stored in a candidate pool for matching with our output. These candidate masks do not contain any textual semantic information. For image $\mathcal{I}$, our constructed candidate mask pool $\mathcal{M}_{SAM}$ is as follows:
\begin{equation}
\mathcal{M}_{SAM}=\left\{\widetilde M_1,\widetilde M_2, \ldots, \widetilde M_K \mid \widetilde M_k \in \mathbb{R}^{H \times W}\right\}.
\vspace{-1em}
\end{equation}
\label{smr}\par
Then we propose two matching score calculation methods to meet for different segmentation error cases. Specially, $s^k_1$ is for addressing under-segmentation cases, while $s^k_2$ is for addressing over-segmentation. We will provide detailed working principles about these two issues in the supplementary materials.
\begin{equation}
s_1^k = \frac{ \sum_{j=1}^{H \times W} \left(\hat{M}_{j}^n \cap \widetilde{M}_{j,k}\right)}{\sum_{j=1}^{H \times W} \left(\hat{M}_{j}^n\right) + \epsilon},
\label{7}
\end{equation} 

\begin{equation}
s_2^k = \frac{ \sum_{j=1}^{H \times W} \left(\hat {M}_{j}^n \cap \widetilde{M}_{j,k}\right)}{\sum_{j=1}^{H \times W}\left(\widetilde{M}_{j,k}\right)},
\label{8}
\vspace{-0.5em}
\end{equation}\\
where $\hat {M}^n_{j}$ and $\widetilde M_{j,k}$ is $j$-th pixel in self-attention enhanced map for phrase $n$ and $k$-th SAM candidate mask, respectively.                         $\epsilon$ is the smoothing factor to prevent a denominator of zero.\par
Then we will select candidate masks with matching scores higher than the threshold $\tau$ to obtain the set $\mathcal{M}^n_{matched}$ as follows:
\begin{equation}
    \mathcal{M}^n_{matched}=\left\{\widetilde{M}_k \mid s_1^k > \tau \;  or \;  s_2^k>\tau  \right\} \in \mathbb{R}^{N_{K} \times H \times W},
\end{equation}
where $\tau$ is matching threshold and $M^{n}_{matched}$ is matched mask pool for corresponding phrase $n$. And $N_k$ is the size of matched mask pool.\par
After matching, if there are SAM candidate masks in the matching set, we will compute the union of these masks to obtain the refined mask $\mathcal{M}^n_{refined}$. If none of the masks generated by SAM can match the masks enhanced by self-attention, then the masks enhanced by self-attention $\hat M^n $ will be considered as the final masks, as described below:
\begin{equation}
\mathcal{M}^n_{refined}=\begin{cases} \cup_{\widetilde{M}_k \in \mathcal{M}^n_{matched}} \widetilde{M}_k \in \mathbb{R}^{H \times W},& if \quad \mathcal{M}^n_{matched} \neq \varnothing \\ \hat M^n ,& else \end{cases}.
\end{equation}
\section{Experiments}
\subsection{Datasets}
\noindent{\bf PNG Dataset.}We conduct experiments on the PNG dataset~\cite{gonzalez2021panoptic}  and compare our method with existing fully-supervised methods. The PNG dataset consists of image-text pairs. Unlike datasets such as RefCOCO~\cite{yu2016modeling}, PNG dataset is characterized by lengthy descriptions of all the objects and their relationships within the complete image. Additionally, the objects within noun phrases encompass both singular and plural forms, making the alignment between visual and textual elements more complex. And PNG dataset comprises 133,103 training images and 8,380 testing images, with 57\% of the noun phrases in the narrative texts belonging to \textit{things} and 43\% to \textit{stuff}.

\subsection{Evaluation Metrics}
Consistent with previous PNG works~\cite{gonzalez2021panoptic,ding2022ppmn,yang2023semi}, we use Average Recall as the evaluation metric, which measures the performance by calculating the Intersection over Union (IoU) between our prediction and the ground truth for each noun phrase, with an IoU~\cite{jiang2018acquisition} threshold. Specifically, we compute the IoU between the predicted masks and bounding boxes generated by our method and the corresponding ground truth. Different thresholds were used to calculate the corresponding recall rates.
Then we compute recall at various IoU thresholds, forming a curve, and the Average Recall refers to the area under this curve. To comprehensively assess the performance, we calculate the Average Recall evaluation metric for \textit{overall}, \textit{things}, \textit{stuff}, \textit{singulars}, and \textit{plurals} splits.

\subsection{Implementation Details}
 Our DiffPNG offers a zero-shot solution, requiring only an inference process, without the need for any training images or annotations. All experiments are conducted on an RTX3090 GPU. All test images are resized to the resolution of 512 × 512.  We employ the pre-trained Stable Diffusion~\cite{rombach2022high} (V1.4) as our generative model and utilize the pre-trained ViT-Huge version for SAM and its automatic mask generation pipeline~\cite{kirillov2023segment} in generating multi-scale masks offline. For stable diffusion, the guidance scale is set to 7.5, the total step is 1000, the DDIM diffusion step is 20, and the DDIM noise schedular $\beta_1,...,\beta_T$ starts from 0.00085 to 0.012 generated by scaled linear.
\subsection{Comparisons with State-of-the-Arts and Alternatives}
\begin{table*}[t]
\centering
\setlength{\tabcolsep}{0.9pt}
\caption{Comparison of our proposed method with the fully-supervised methods and zero-shot methods on the PNG benchmark.}
\vspace{-10pt}
\resizebox{0.86\textwidth}{!}{
\begin{tabular}{@{}c|c|c|c|ccccccc@{}}
\toprule

\multicolumn{2}{c|}{\multirow{2}{*}{Method}} &Visual&Language&\multicolumn{5}{c}{Segmentation Average Recall(\%) $\uparrow$}\\
 \multicolumn{2}{c|}{}&Encoder&Model&Overall & Singular & Plural & Thing & Stuff \\

\midrule
\multirow{9}{1.5cm}{Fully-Supervised} &PNG~\cite{gonzalez2021panoptic}&RN-101&BERT&55.4&56.2&48.8&56.2&54.3\\ 
&PPMN~\cite{ding2022ppmn}&RN-101&BERT&59.4&60.0&54.0&57.2&62.5\\
&MCN~\cite{luo2020multi}&RN-101&GRU&54.2&56.6&38.8&48.6&61.4\\
&EPNG~\cite{wang2023towards}&RN-101&BERT&58.0&58.6&52.1&54.8&62.4\\
&DRMN~\cite{lin2023context}&RN-101&BERT&62.9&63.6&56.7&60.3&66.4\\
&PPO-TD~\cite{hui2023enriching}&RN-50&BERT&66.1&69.1&58.3&64.0&70.7\\
&PiGLET~\cite{gonzalez2023piglet}&SWIN-L&BERT&65.9&67.2&54.5&64.0&68.6\\
&NICE~\cite{wang2023nice}&RN-101&BERT&62.3&63.1&55.2&60.2&65.3\\
&XPNG~\cite{guo2024improving}&RN-101&BERT&63.3&64.0&56.4&61.1&66.2\\
\midrule
\multirow{5}{1.5cm}{Zero-Shot} &Random&-&-&15.53&15.76&13.47&11.90&20.58\\
&DatasetDiffusion~\cite{nguyen2024dataset}&UNET&CLIP&23.46&24.28&16.03&15.99&33.83\\
&DiffSeg~\cite{tian2023diffuse}&UNET&CLIP&24.08&24.75&18.02&17.69&32.95\\
\cmidrule{2-9}
&DiffPNG(ours)&UNET&CLIP&33.54       & 34.05      & 28.93      & 29.79     &38.74\\
&DiffPNG*(ours)&UNET&CLIP&\textbf{38.49}  &\textbf{39.18}  & \textbf{32.14} & \textbf{35.95} &  \textbf{42.00}\\
\bottomrule
\end{tabular}}
\vspace{-1.5em}
\label{tab:1}
\end{table*}

As shown in Tab.~\ref{tab:1}, we evaluate the performance of our proposed method, DiffPNG, along with fully-supervised methods and different zero-shot methods on the PNG dataset.
Fully-supervised methods achieved good performance at the expense of training costs. Under the zero-shot setting, our DiffPNG method has achieved the best performance, significantly surpassing other zero-shot methods based on the diffusion model and zero-shot learning. Compared to Dataset Diffusion~\cite{nguyen2024dataset} and DiffSeg~\cite{tian2023diffuse}, DiffPNG achieved 33.54\% on the Average
Recall evaluation metric, improving by +10.08\% and +9.46\% respectively. 
Under the zero-shot setting where masks are randomly generated, the performance is merely 15.53\%, which underscores the effectiveness of the diffusion-based paradigm we've crafted in achieving robust image-text alignment capabilities. Additionally, DiffPNG has further improved performance through an enhanced version called DiffPNG* (with an enhanced SMR module). DiffPNG* achieves an overall Average Recall of 38.49\%, with a +4.95\% improvement over the original DiffPNG.

\subsection{Ablation Studies}
\begin{table*}[ht]
\centering
\setlength{\tabcolsep}{0.7pt}
\setlength{\tabcolsep}{4pt}
\caption{Ablation study of LSP Module.}
\vspace{-1em}
\begin{tabular}{@{}c|c|c|ccccc@{}}
\toprule
\multicolumn{3}{c|}{\multirow{2}{*}{}} & \multicolumn{5}{c}{Segmentation Average Recall(\%) $\uparrow$}\\
\multicolumn{3}{c|}{\multirow{2}{*}{}}&Overall & Singular & Plural & Thing & Stuff \\
\midrule
    \multicolumn{3}{c|}{Baseline(only use cross attention)} &30.24 & 30.65 & 26.58 & 26.88 & 34.92 \\
\midrule
\multirow{2}{*}{(a)Alternatives} &\multicolumn{2}{c|}{DatasetDiffusion~\cite{nguyen2024dataset}} &23.46&24.28&16.03&15.99&33.83 \\
\multirow{3}{*}{} &\multicolumn{2}{c|}{DiffSeg~\cite{tian2023diffuse}}  &24.08&24.75&18.02&17.69&32.95\\
\midrule
\multirow{8}{*}{(b)Selection} & \multirow{4}{*}{Percentage $K$} 
& 90 &  29.35     &  29.83     &   25.01    &    26.23   & 33.68 \\
&       & 80    & 30.13      &   30.61    &  25.79     & 25.99      &  35.88\\
&       & 70   &  28.81     &  29.27     &  24.67     &   24.12    &35.34  \\
&       & 60   &    29.35   &    29.83   &  25.01     &  26.23     &  33.68
\\
\cmidrule{2-8}
 & \multirow{4}{*}{Threshold $\beta$} & 0.3 & 32.04      &  32.54     & 27.47      &28.05       & 37.58 \\

&       &  \cellcolor{gray!18} 0.4   & \cellcolor{gray!18}\textbf{33.54 }      & \cellcolor{gray!18}\textbf{34.05 }     & \cellcolor{gray!18}\textbf{28.93}      &\cellcolor{gray!18} \textbf{29.79 }    & \cellcolor{gray!18} \textbf{38.74} \\
&       & 0.5   & 32.81      &   33.32    &   28.13    &  29.48     & 37.43 \\
&       & 0.6   &  30.15     &   30.67    &   25.45    &  27.30     &34.11  \\

\midrule
&\textbf{cross}&\textbf{self} &Overall & Singular & Plural & Thing & Stuff \\
\cmidrule{2-8}
\multirow{7}{*}{(c)Resolution} & \multirow{2}{*}{16} &  \cellcolor{gray!18} 32    &\cellcolor{gray!18} \textbf{33.54}       & \cellcolor{gray!18} \textbf{34.05}      & \cellcolor{gray!18}\textbf{28.93 }     &\cellcolor{gray!18} \textbf{29.79}      &\cellcolor{gray!18} \textbf{38.74}  \\
&       & 64    & 33.45 & 33.96 & 28.80  & 29.79 & 38.53 \\
\cmidrule{2-8}
& \multirow{2}{*}{32} & 32    & 16.54      & 15.98     &  21.57     &   21.15    & 10.13 \\
&       & 64    & 16.86      &  16.29     &  22.01     &    21.43   & 10.51 \\
\cmidrule{2-8}
& \multirow{2}{*}{64} & 32    & 11.68      &11.73       &  11.28     &  9.69     & 14.44 \\
&       & 64    &   11.09    & 11.05      &    11.48   &  10.15     & 12.40  \\
\bottomrule
\end{tabular}
\vspace{-0.5em}
\label{tab:csai}

\end{table*}

\noindent{\bf Locate-to-Segment Processor.} In our ablation study of the LSP module, as depicted in Tab.~\ref{tab:csai}, we focus on validating the effectiveness of our approach, which identifies anchor points using cross-attention mechanisms and subsequently performs segmentation with self-attention. In the first row, cross-attention alone serves as our baseline. As shown in Tab.~\ref{tab:csai}(a), we compare the methods of solely employing cross-attention versus combining it with self-attention in pure visual segmentation tasks. DatasetDiffusion~\cite{nguyen2024dataset} entails taking the power of self-attention and multiplying it with cross-attention, while DiffSeg~\cite{tian2023diffuse} involves assigning semantics based on the merged self-attention and cross-attention. Based on the experimental results, neither of these two methods appears to be well-suited for the PNG dataset, as they both show a decrease in performance compared to the baseline.
In Tab.~\ref{tab:csai}(b), we refine the baseline for utilizing cross-attention to select anchor pixels by adjusting the way and parameter. Initially, we devise a method where the top K confidence-ranked points on cross-attention are chosen as anchor pixels. However, this approach does not yield performance improvements. We attribute this to the inflexibility of the percentile method, which selects a fixed number of points for all samples. Consequently, we adopt a simpler and more adaptable threshold-based approach, selecting pixels with confidence scores above a threshold $\beta$ as anchor points. Upon fine-tuning the threshold values, we observe a significant improvement of +3.30\% than baseline when $\beta$ was set to 0.4. Finally, in Tab.~\ref{tab:csai}(c), we conduct an ablation study on the impact of the resolution sizes of cross-attention and self-attention on experimental performance. Ultimately, we find setting the cross-attention resolution to 16 and the self-attention resolution to 32 yields the best results for DiffPNG.\\

\begin{table*}[t]
\centering
\caption{Ablation study of different feature aggregators for noun phrases.}
\vspace{-5pt}
\resizebox{0.86\textwidth}{!}{
    \begin{tabular}{@{}c|c|ccccc@{}}
    \toprule
\multicolumn{2}{c|}{\multirow{2}{*}{}} & \multicolumn{5}{c}{Segmentation Average Recall(\%) $\uparrow$}\\
\multicolumn{2}{c|}{\multirow{2}{*}{}}&Overall & Singular & Plural & Thing & Stuff \\
     \midrule
    \multirow{3}{*}{(a)Aggregator}&Average aggregator& 32.40 & 33.11 & 25.91 & 28.80 & 37.39\\
    &Multiplication aggregator & 28.67 &30.75 & 19.92 & 25.86 & 34.97 \\
    &Subject-Focused aggregator &33.27&33.81&28.41&29.62&38.35 \\
    \midrule
     \multirow{2}{*}{(b)Input feature}& self-attention aggregation&33.27&33.81&28.41&29.62&38.35 \\
    &\cellcolor{gray!18}cross-attention aggregation  &\cellcolor{gray!18} \textbf{33.54}       & \cellcolor{gray!18} \textbf{34.05}      & \cellcolor{gray!18} \textbf{28.93 }     & \cellcolor{gray!18} \textbf{29.79}      & \cellcolor{gray!18}\textbf{38.74}  \\
\bottomrule
\end{tabular}
}
\vspace{-1.5em}
\label{tab:sffa}
\end{table*}

\begin{table*}[t]
\centering
\setlength{\tabcolsep}{7pt}
\caption{Ablation study of SMR Module.}
\vspace{-6pt}

\resizebox{0.88\textwidth}{!}{
    \begin{tabular}{@{}cc|ccccc@{}}
    \toprule
     \multicolumn{2}{c|}{\multirow{2}{*}{}} & \multicolumn{5}{c}{Segmentation Average Recall(\%) $\uparrow$}\\
 \multicolumn{2}{c|}{}&Overall & Singular & Plural & Thing & Stuff \\
    \midrule
    \multicolumn{2}{c|}{Baseline} & 30.24 & 30.65 & 26.58 & 26.88 & 34.92 \\
    \multicolumn{2}{c|}{DiffPNG} &\textbf{33.54}       & \textbf{34.05}      & \textbf{28.93 }     & \textbf{29.79}      &\textbf{38.74}  \\
    \midrule
    \multicolumn{2}{c|}{+DenseCRF} & 23.49& 24.13 & 27.71 & 20.97 & 26.69 \\
    \midrule
    \multicolumn{2}{c|}{+SMR(DiffPNG*)} &\textbf{38.49}  &\textbf{39.18}  & \textbf{32.14} & \textbf{35.95} &  \textbf{42.00}\\
    \midrule
    \multirow{5}{3cm}{\centering Threshold $\tau$} & 
    0.3   & 36.35 & 36.92 & 31.18 & 33.38 & 40.49 \\
    & 0.4   &37.55  & 38.16 & 32.02 & 34.83 &41.33  \\
    & 0.5  &38.26&38.92&32.28&35.70&41.82 \\
     &\cellcolor{gray!18} 0.6   &\cellcolor{gray!18}\textbf{38.49}  &\cellcolor{gray!18}\textbf{39.18}  &\cellcolor{gray!18} \textbf{32.14} &\cellcolor{gray!18} \textbf{35.95} &\cellcolor{gray!18}  \textbf{42.00}\\
          & 0.7   &38.23  & 38.96 & 31.68 & 35.57 & 41.93 \\
\bottomrule
\end{tabular}}
\vspace{-2em}
\label{tab:refine}
\end{table*}

\noindent{\bf Subject-Focused Feature Aggregator.}
In Tab.~\ref{tab:sffa}, we investigate the impact of different feature aggregators on Average
Recall, as well as different input features into the aggregator on experimental outputs. The results indicate that our proposed aggregator, which based on semantic similarity, surpasses both the average aggregator $\mathcal{A}^n_{fuse} =\frac{1}{N_p} \sum^{N_p}_{i=1}\mathcal{A}^{n,i}_{enhanced}$ and multiplication aggregator $\mathcal{A}^n_{fuse} = \prod^{N_p}_{i=1}\mathcal{A}^{n,i}_{enhanced}$, with 33.27\% overall Average Recall. Moreover, aggregating features through cross-attention achieves +0.27\% improvement on overall Average Recall over aggregating features after enhancing with self-attention.\par
\noindent{\bf SAM-based Mask Refinement.} 
In Tab.~\ref{tab:refine}, we conduct experiments on our designed post-processing refinement SMR module to verify the effectiveness and perform hyper-parameter ablation. The first row and the second row represent the results only using the cross-attention baseline and incorporating our designed LSP module, respectively. In the third row, we add the traditional refinement method DenseCRF~\cite{krahenbuhl2011efficient} (based on local relationships defined by color and distance of pixels). The results show that the addition of DenseCRF does not lead to performance improvement, which may be attributed to the coarseness of the masks in a zero-shot manner. Without carefully designed refinement modules, it is easy to introduce noise. In the fourth row, we add our designed SMR module, which achieves the best performance. We perform ablation on the threshold $\tau$ and find that setting it to 0.6 yields the best results. The overall Average Recall reaches 38.49\%, which is a +4.95\% improvement over the original DiffPNG and a +8.25\% improvement over the baseline. This demonstrates the effectiveness of the SMR module in refining segmentation masks.

\begin{table*}[t]
    \centering
    \caption{Ablation study on different components.}
    \vspace{-1em}
    \begin{tabular}{@{}c|ccc|ccccccc@{}}
    \toprule
    \multirow{2}{*}{ID} &\multirow{2}{*}{CA} &\multirow{2}{*}{LSP} & \multirow{2}{*}{SMR} &\multicolumn{5}{c}{Segmentation Average Recall(\%) $\uparrow$}\\
     &&&&Overall & Singular & Plural & Thing & Stuff \\ 
    \midrule
    (a)&&&& 15.53 & 15.76 & 13.47 & 11.9 & 20.58\\
    (b)&\cmark&&& 30.24 & 30.65 & 26.58 & 26.88 & 34.92\\
    (c)&\cmark&\cmark &&33.54       & 34.05      & 28.93      & 29.79     &38.74 \\
    \rowcolor{gray!18}(d)&\cmark& \cmark &\cmark &\textbf{38.49}  &\textbf{39.18}  & \textbf{32.14} & \textbf{35.95} &  \textbf{42.00} \\ \bottomrule
    \end{tabular}
    \vspace{-1em}
    \label{tab:ablation}
\end{table*}

\noindent{\bf Effect of different components.} In Tab.~\ref{tab:ablation}, we summarize the ablation study of each module proposal, comparing the impact of different component combinations on the performance. The row(a) represents results generated randomly. The row(b) indicates a significant improvement of +14.71\% in overall Average
Recall compared to random configuration after incorporating only the cross-attention component. This suggests that the diffusion model possesses a certain level of image-text alignment capability in the discrimination task. The row(c) demonstrates a further enhancement of +3.30\% in overall Average
Recall after adding the LSP component, indicating an improvement over the previous configuration and validating the effectiveness of our proposed paradigm of localization before segmentation. Leveraging self-attention effectively boosts the method's performance. Lastly, we add the SMR component in row(d), refining masks using SAM, resulting in the best performance. The overall Average
Recall reaches 38.49\%, a significant improvement of +22.96\% compared to using the random configuration.
\begin{figure*}[t] 
\centering 
\includegraphics[width=1\columnwidth]{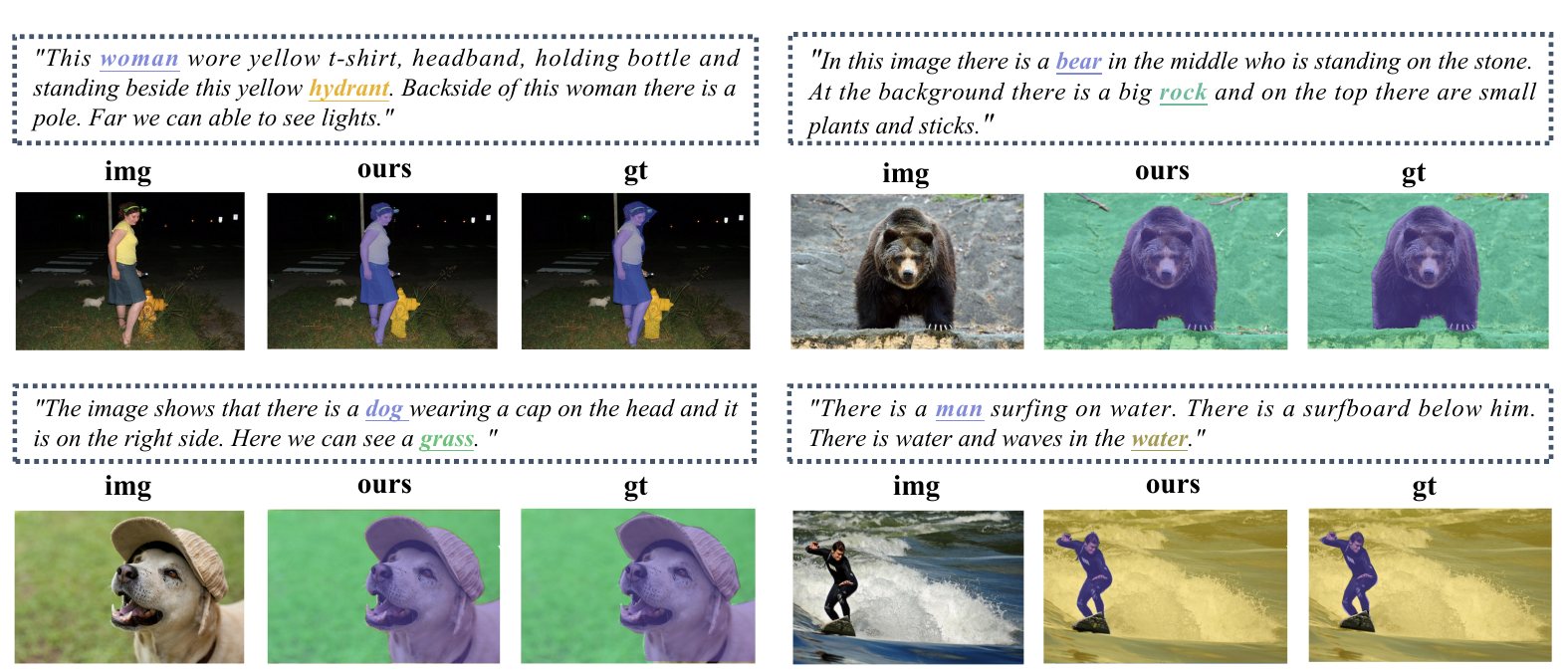}  
\vspace{-1.8em}
\caption{
Qualitative analysis compares our proposed DiffPNG* with ground truth.} 
\label{fig:qualitative}
\vspace{-1.5em}
\end{figure*}
\subsection{Qualitative Results}
\label{sect:qualitative results}
As shown in Fig.~\ref{fig:qualitative}, we visualize our proposed DiffPNG* compared with ground truth on selected samples from the PNG benchmark dataset. It can be observed that our solution, based on the diffusion model, achieves excellent zero-shot inference capability, even without having seen any data from the PNG dataset. The results demonstrate that our proposed paradigm can achieve phrase perception. We will give more visualizations of each module's effectiveness in the supplementary materials.
\section{Conclusion}
In this paper, we reformulate the Panoptic Narrative Grounding task using stable diffusion as a zero-shot problem of initial localization, followed by segmentation, and finally refinement. By leveraging the reverse diffusion process, we derive cross-attention and self-attention maps to construct our Locate-to-Segment Processor module. Furthermore, by harnessing SAM's multi-scale capability to further refine our masks, we enhance mask quality, showcasing our framework's proficiency in understanding phrase-level text-image alignment. Experimental results demonstrate that DiffPNG can significantly improve segmentation performance, suggesting that employing generative models for discriminative tasks such as noun phrase understanding holds promise.

\subsubsection{Acknowledgements.}
This work was supported by National Science and Technology Major Project (No. 2022ZD0118201), the National Science Fund for Distinguished Young Scholars (No.62025603), theNational Natural Science Foundation of China (No.U21B2037, No.U22B2051, NO62072389), the National Natural Science Fund for Young Scholars of China (No.62302411), China Postdoctoral Science Foundation (No.2023M732948), the NaturalScience Foundation of Fujian Province of China (No.2021J01002, No.2022J06001), andpartially sponsored by CCF-NetEase ThunderFire lnnovation Research Funding (NO.CCF-Netease 202301).

%
%
\bibliographystyle{splncs04}
\bibliography{main}

\clearpage

\appendix
\renewcommand{\thesection}{\Alph{section}}
\section{Appendix}

\subsection{The Error Accumulation in Multi-step Approaches}
\begin{figure}[h]
    \centering
    \vspace{-1em}
\includegraphics[width=0.9\columnwidth]{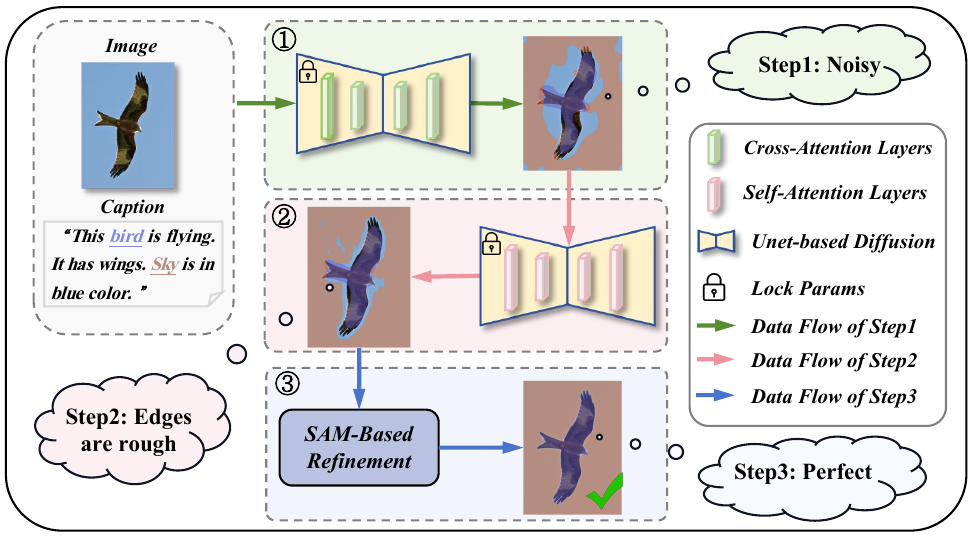}
    \caption{An example demonstrates how our model mitigated the issue of error accumulation in this multi-step approach.}
    \label{fig:error_accumulation}
    \vspace{-2em}
\end{figure}

Our multi-step approach is designed as a refinement process from coarse to fine to address error accumulation. The Fig.~\ref{fig:error_accumulation} illustrates this process using real examples. First, we use visual-text correlation in cross-attention to locate text-related instances. Next, self-attention is employed to obtain a more accurate mask through its self-correlation. Finally, SAM is used to refine the results.
\subsection{The Working Principles of SMR}
\begin{figure}[h]
  \centering 
  \includegraphics[width=0.9\linewidth]{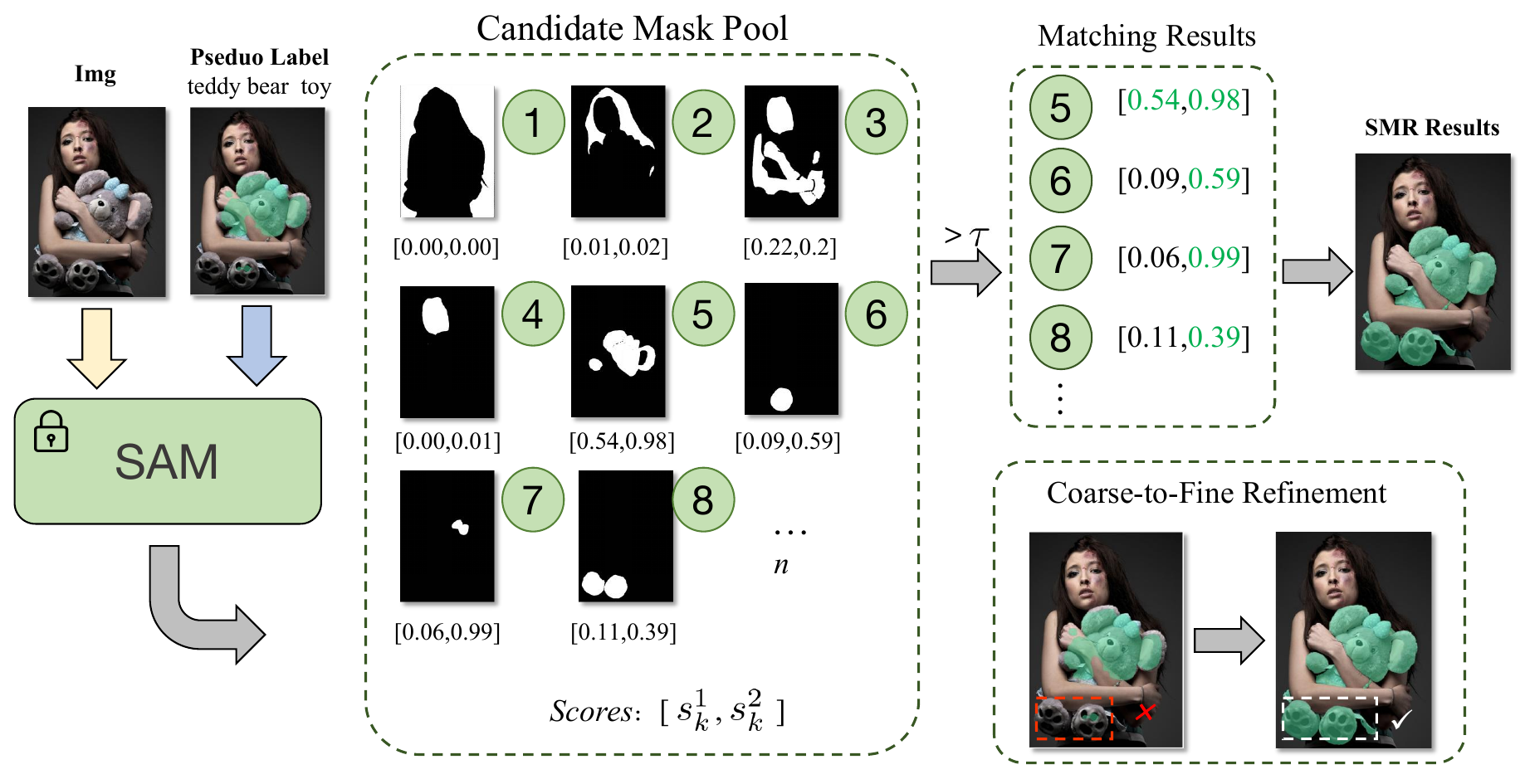} 
  \caption{An example illustrating the working principle of the SAM-based Mask Refinement (SMR) strategy.}
  \label{fig:principal}
\end{figure}

To enhance clarity regarding the working principles underlying the SAM-based Mask Refinement (SMR) strategy introduced in our work, we provide a comprehensive explanation, as depicted in Fig.~\ref{fig:principal}. Specifically, we select an example to elucidate how SMR refines the coarse mask to a better refined one. SMR utilizes multi-scale features obtained from the SAM automatic pipeline, systematically selects partial segments with matching scores exceeding a certain threshold, and merges them together to form the final result. Due to the under-segmentation and over-segmentation issues present in the coarse segmentation results, we propose two matching score calculation formulas, both $s^k_1$ and $s^k_2$ compute the intersection of the prediction and each SAM candidate mask. When the area of the predicted foreground is smaller than the area of the candidate mask, by retaining candidates with a score greater than $s^k_1$, we can correct under-segmentation in predictions. Similarly, we need $s^k_2$ to tackle the case that the area of prediction is larger than the proposal mask. As shown in Fig.~\ref{fig:principal}, for this teddy bear toy example, the prediction mask suffers from under-segmentation. Our matching scheme effectively matches and combines SAM's candidate masks to form a complete mask.
\subsection{Visualization Showcasing SMR Addresses Under-segmentation and Over-segmentation Issues}
\begin{figure}[h]
    \centering
    \vspace{-1em}
\includegraphics[width=0.9\columnwidth]{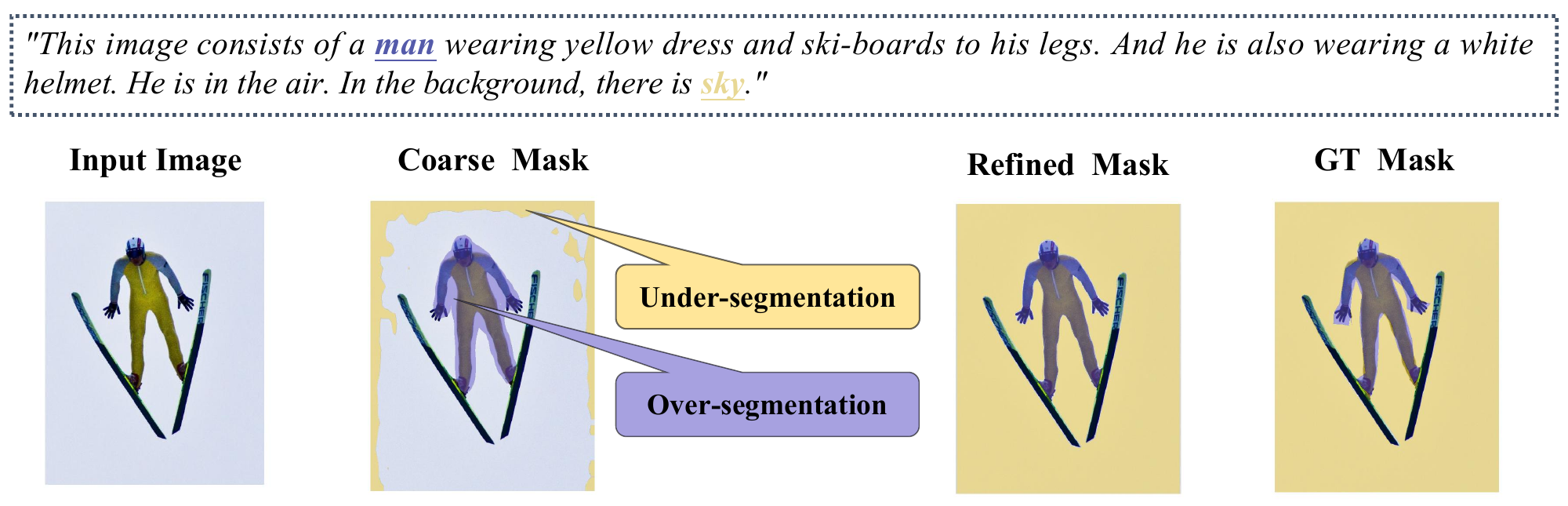}
    \caption{Visualization showcasing how the SMR module effectively addresses both under-segmentation and over-segmentation issues.}
    \label{fig:under_over}
    \vspace{-2em}
\end{figure}

After introducing the principles of SMR, we showcase an example to illustrate how our SMR module effectively addresses both under-segmentation and over-segmentation issues, as is shown in Fig.~\ref{fig:under_over}. In this example, the prediction of the sky exhibits under-segmentation, with some pixels mismatched. Conversely, the man displays over-segmentation at the edge. Surprisingly, our designed SMR module effectively addresses both of these issues.

\subsection{Ablation of Binarization Threshold} 
Our zero-shot framework is based on cross-attention and self-attention maps, resulting in a probability distribution where each value on the maps represents a probability between 0 and 1. We need to undergo binarization to convert it into a binary mask, with values of either 0 or 1. We conduct experiments about the binarization threshold $\alpha$ in Tab.~\ref{tab:s2h}. We investigate it by using the output of the Locate-to-Segment Processor~(LSP) module and observe the impact of different thresholds on overall Average Recall. The results indicate that setting the threshold $\alpha$ to \textbf{0.3} achieves optimal performance, with an overall average recall rate reaching \textbf{33.54\%}, while the performance at other threshold settings is relatively lower. This suggests that choosing an appropriate threshold $\alpha$ is a critical strategy in binarization threshold settings.
\begin{table}[h]
\centering
\vspace{-1em}
\setlength{\tabcolsep}{4pt}
\caption{Ablation of different thresholds which convert a probability to a binary mask. We conduct this ablation study on the output of the LSP module.}
\resizebox{0.6\textwidth}{!}{
    \begin{tabular}{c|ccccc}
    \toprule
     \multirow{2}{*}{$\alpha$}& \multicolumn{5}{c}{Segmentation Average Recall(\%) $\uparrow$}\\
 &Overall & Singular & Plural & Thing & Stuff \\
    \midrule
0.2&32.96  &33.55 & 27.63& 28.31 &\textbf{39.41} \\
          \rowcolor{gray!18} 0.3  & \textbf{33.54}       & \textbf{34.05}      & \textbf{28.93 }     & \textbf{29.79}      &38.74  \\
          0.4 & 32.14 &32.56 &28.36 & 29.53 & 35.77 \\
          0.5 &28.20  & 28.52& 25.28& 27.09 & 29.74 \\
\bottomrule
\end{tabular}
}
\vspace{-2em}
\label{tab:s2h}
\end{table}



\subsection{The Visualization of Effectiveness for Each Module} 
\begin{figure}[h]
    \centering
    \vspace{-1em}
    \includegraphics[width=1.0\columnwidth]{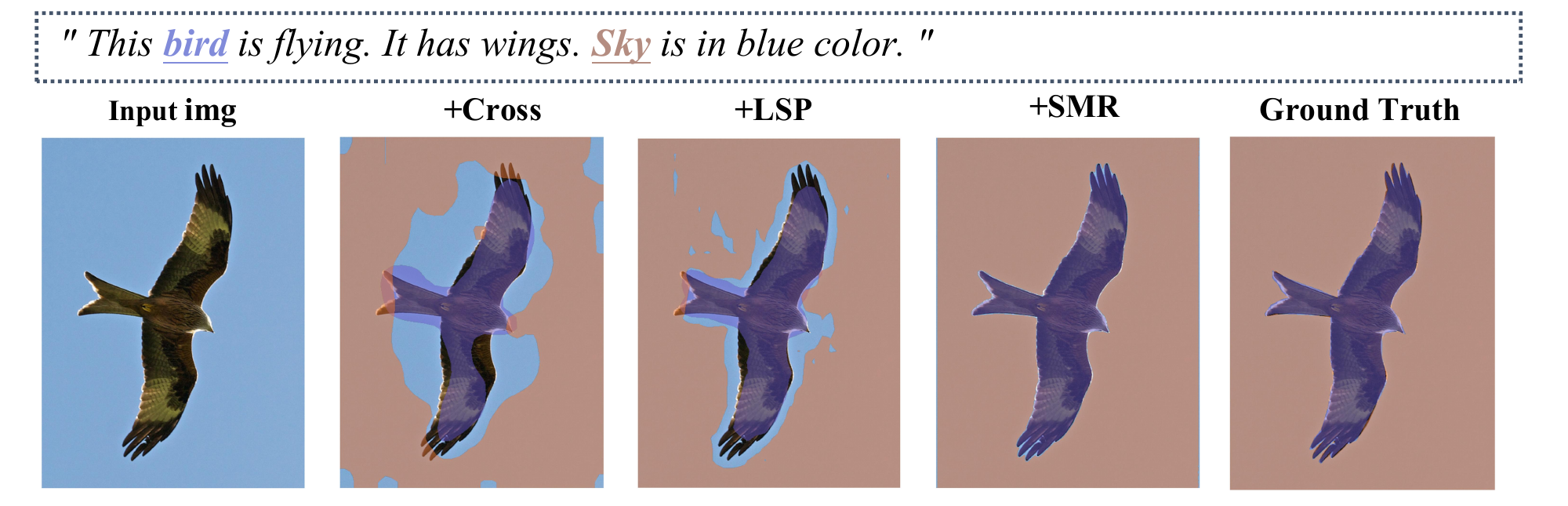}
    \caption{The visualization of effectiveness for each module. Notably, each module is experimented based on the output of the preceding module.}
    \label{fig:ablation_vis}
    \vspace{-1em}
\end{figure}

We visualize the intermediate and final results of our proposed framework, as illustrated in Fig.~\ref{fig:ablation_vis}. It can be observed that the results obtained solely from the cross-attention map are relatively coarse. However, the LSP, which selects anchor points from the cross-attention map and then aggregates self-attention map, further refines the masks. Ultimately, with the enhancement from SMR, the results are comparable to the ground-truth.
\subsection{More Comparative Visualizations of Our Method} 
\begin{figure}[h]
    \centering
    \vspace{-5em}
    \includegraphics[width=1.05\columnwidth]{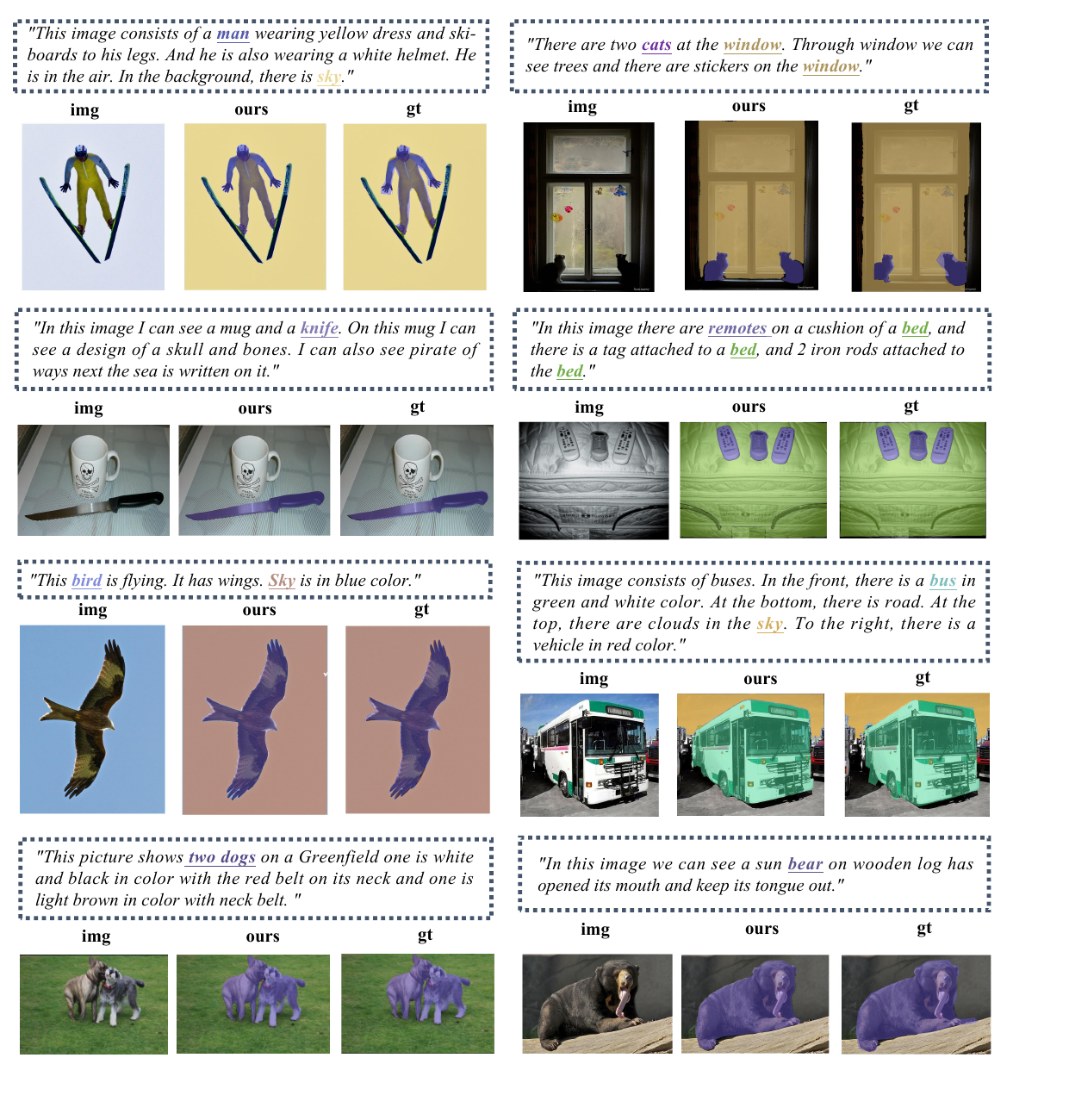}
    \caption{More visualization results of our proposed DiffPNG*, compared with the ground-truth. }
    \label{fig:vis_more}
\end{figure}

As Fig.~\ref{fig:vis_more} shows, we present more comparative visualizations of our proposed method with ground-truth. It can be observed that our zero-shot approach based on the diffusion model achieves promising results.
\end{document}